\documentclass[sigconf]{acmart}
\usepackage{multirow}
\usepackage{algorithm}
\usepackage{algorithmic}
\AtBeginDocument{%
  }

\copyrightyear{2026}
\acmYear{2026}
\setcopyright{cc}
\setcctype{by}
\acmConference[WWW '26]{Proceedings of the ACM Web Conference 2026}{April 13--17, 2026}{Dubai, United Arab Emirates}
\acmBooktitle{Proceedings of the ACM Web Conference 2026 (WWW '26), April 13--17, 2026, Dubai, United Arab Emirates}
\acmPrice{}
\acmDOI{10.1145/3774904.3792488}
\acmISBN{979-8-4007-2307-0/2026/04}

\usepackage[shortcuts]{extdash}

\begin{document}

\title[Predictability-Aware Compression and Decompression Framework for Multichannel Time Series \\ Data with Latent Seasonality]{Predictability-Aware Compression and Decompression Framework for Multichannel Time Series Data with Latent Seasonality}

\author{Ziqi Liu}
\affiliation{%
  \department{Department of Computer Science}
  \institution{The University of Texas at Dallas}
  \city{Dallas}
  \state{TX}
  \country{USA}
}
\email{ziqi.liu@utdallas.edu}

\author{Pei Zeng}\affiliation{%
    \department{Global AI \& Platforms}
    \institution{LexisNexis}
    \city{Raleigh}
    \state{NC}
    \country{USA}}
    \email{pei.zeng@lexisnexis.com}

\author{Yi Ding}
\affiliation{%
  \department{Department of Computer Science}
  \institution{The University of Texas at Dallas}
  \city{Dallas}
  \state{TX}
  \country{USA}
}
\email{yi.ding@utdallas.edu}

\begin{abstract}
Real-world multichannel time series prediction faces growing demands for efficiency across edge and cloud environments, making channel compression a timely and essential problem. Motivated by success of Multiple-Input Multiple-Output (MIMO) methods in signal processing, we propose a predictability-aware compression–decompression framework to reduce runtime, decrease communication cost, and maintain prediction accuracy across diverse predictors. The core idea involves using a circular seasonal key matrix with orthogonality to capture underlying time series predictability during compression and to mitigate reconstruction errors during decompression by introducing more realistic data assumptions. Theoretical analyses show that the proposed framework is both time-efficient and accuracy-preserving under a large number of channels. Extensive experiments on six datasets across various predictors demonstrate that the proposed method achieves superior overall performance by jointly considering prediction accuracy and runtime, while maintaining strong compatibility with diverse predictors.
\end{abstract}





\maketitle

\section{Introduction}


Multichannel time series prediction is widely studied across various domains \cite{Intro19, Intro20, Intro21, Intro22} and has become increasingly critical in Web-scale applications, where time series data is transmitted through Internet of Things (IoT) networks to Web platforms for large-scale access and decision support.
In this context, a channel refers to a single signal stream or time-dependent variable—such as the reading from one sensor, a specific feature dimension, or a data source—recorded in parallel with other channels as part of a multichannel time series.
These applications typically stream data from sensors to Web platforms, posing practical challenges for transmission and computation. For example, in large-scale IoT deployments, low-cost sensors rely on edge devices to aggregate local data and transfer it from Low Power Wide Area (LPWA) networks to IP link network \cite{Intro29}. In this case, as shown in Figure \ref{fig:Intro1}, multichannel data may originate from a single sensor \cite{Intro28} or from edge devices aggregating multiple sensors in areas without IP coverage \cite{Intro30, Intro31}, both of which strain bandwidth and computation. Therefore, time series data compression for prediction task is a timely and valuable research direction.


{
\setlength{\intextsep}{1pt}
\begin{figure}[ht]
    \centering
    \includegraphics[width=\linewidth]{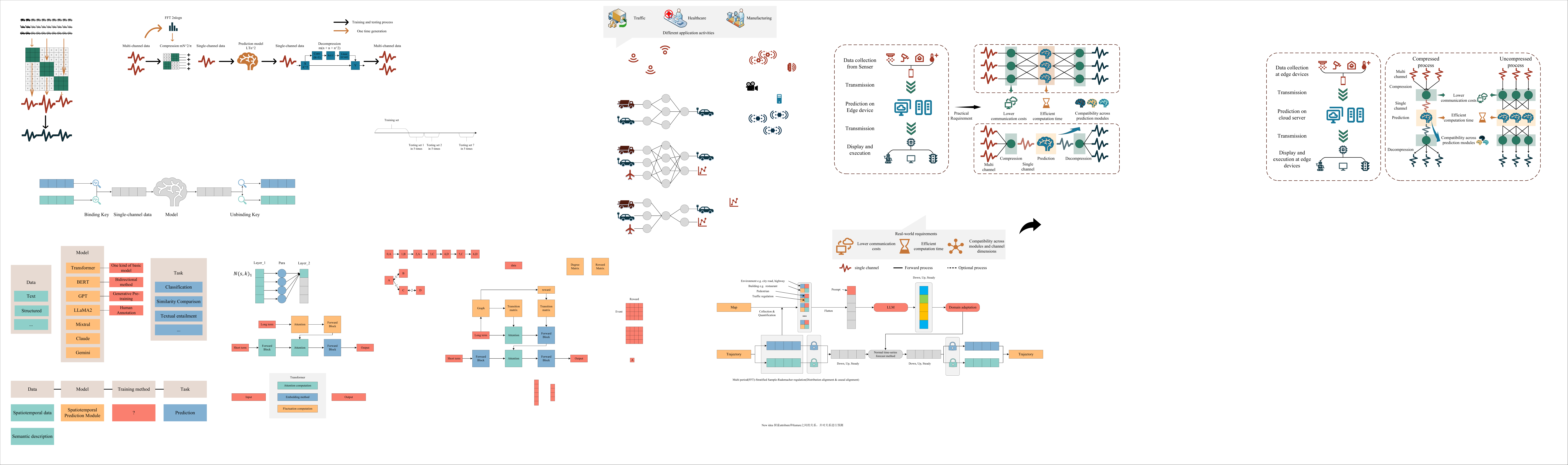}
    \vspace{-20pt}
    \caption{Time series data compression in two example situations.}
    \label{fig:Intro1}
    \Description{None.}
\end{figure}
}

Existing time series data compression methods primarily focus on storage and computational efficiency through temporal reduction, which can be categorized into three types. (1) Representation-based methods \cite{RW1,RW2,RW3} use mathematical functions or learned nonlinear mappings to obtain compact latent representations. (2) Dictionary-based methods \cite{RW6,RW7} exploit structural redundancy via similarity measures or codebook matching. (3) Sequential algorithms \cite{RW4,RW5,RW6,RW7} process data incrementally using information-theoretic principles for online or model-free prediction. While compressing along the temporal dimension reduce input size, it can obscure crucial short-term patterns and make down-sampling a potentially risky operation for prediction task \cite{Intro25}.

On the other hand, general-purpose dimension-reduction techniques can be directly applied for channel compression, including Principal Component Analysis (PCA) \cite{Intro1}, Multi-Dimensional Scaling (MDS) \cite{Intro2}, Isomap \cite{Intro3}, t-distributed Stochastic Neighbor Embedding (t-SNE) \cite{Intro4}, and Uniform Manifold Approximation and Projection (UMAP) \cite{Intro5}. However, while these methods aim to either capture temporal inter-channel dependencies or perform intra-channel reduction, they are not specifically designed for time series prediction. As a result, they fail to integrate both inter- and intra-channel information in a unified manner, and do not preserve or reveal the underlying predictability of the data after compression. Additionally, many of these methods, originally developed for data visualization or non-regression tasks, are unable to accurately reconstruct the original multichannel structure from the predicted compressed representation.



The Multiple-Input Multiple-Output (MIMO) paradigm in signal processing and the recent success of MIMONet \cite{Intro18} in image compression bring new opportunities to build a compression\-/decompression framework for multichannel time series prediction. However, applying this idea in practice still poses significant challenges.
\textbf{First}, preserving 
the potential predictive information is non-trivial in the compression process. The temporal dependence inherent in time series data stores predictive information and provides a foundation for accurate forecasting. However, such predictability cannot be guaranteed during compression when using current approaches, such as circular convolution with keys sampled from a normal distribution (as in HRR \cite{Intro17}) or a Bernoulli distribution (as in MIMONet).
\textbf{Second}, oversimplified assumptions on data pose a critical challenge for accurate decompression. In particular, the assumption of normally distributed input—adopted in HRR—does not align well with the properties of real-world time series data. This mismatch can result in inadequate decompression modules and additional errors, such as intra-channel interference across timestamps and inter-channel interference within the same timestamp.
\textbf{Third}, the overhead introduced by the compression–decompression processes may offset its efficiency benefits. While compressed data naturally leads to reduced computational and transmission time, the compression-decompression steps themselves may introduce additional overhead, potentially increasing overall time complexity.

To address these challenges, we first theoretically show that redundancy in multichannel time series can be exploited to design a compression method that preserves predictive information. Based on this insight, we propose a predictability-aware compression\-/decompression framework for multichannel time series prediction. 
The predictability is maintained by extracting the underlying seasonal patterns through Fourier transform. In this framework, the compression module uses a circular-key for single channel encoding and multichannel predictability alignment, while the decompression module applies the same circular-key matrix with orthogonality constraints and a residual block to mitigate reconstruction errors. Time-complexity analysis confirms its superior efficiency over direct multichannel predictors. Extensive experiments on six datasets and four backbone predictors demonstrate runtime reduction without sacrificing accuracy, even for data without strong periodicity (e.g., sensor drift data). A Cobb–Douglas-based performance index~\cite{Exper1} jointly evaluates accuracy and efficiency, highlighting our framework’s overall advantage and broad compatibility across diverse predictors.

Our main contributions are summarized as follows:
\begin{itemize}
\item \textbf{Observations and Theoretical Foundations:} We provide theoretical analyses that prove the predictability-awareness of our compression module, grounding the results in empirical observations and supporting analysis.
\item \textbf{Compression–Decompression Framework:} We propose a compression–decompression framework for multichannel time-series prediction that minimizes reconstruction error during decompression. Time-complexity analysis shows that the framework achieves better time efficiency compared with a direct multichannel predictor.
\item \textbf{Experimental Analysis:} We conduct extensive experiments demonstrating that the proposed framework achieves lower runtime and superior performance when jointly considering prediction accuracy and computational cost, while maintaining strong compatibility across diverse module architectures.
\end{itemize}

The following sections are organized into Related Work, Problem Definition, Observations and Theoretical Foundations, Compression and Decompression Framework, and Experiments. The Observations and Theoretical Foundations section shows how our method preserves predictability. The Compression and Decompression Framework section explains the details shown in Fig.~\ref{fig:Method1} and describes overall process in detail.

\section{Related Works}

\paragraph{Time Series Prediction}

Deep learning architectures for multichannel time series forecasting can be broadly categorized into several paradigms. Channel-independent models \cite{Intro9,Intro10} treat each channel separately, ignoring inter-series correlations to achieve scalability and simplicity. In contrast, channel-mixing models \cite{Intro11,Intro12,Intro13} explicitly model dependencies across different channel to improve predictive accuracy. Single-channel architectures \cite{Intro14} are originally designed for single channel inputs but can be generalized to multichannel settings. In addition, recent efforts have explored using large pre-trained models for time series forecasting. Time-LLM \cite{Intro26} reprograms frozen large language model (LLM) by encoding temporal data as token sequences to enable zero-shot forecasting. UniST \cite{Intro27} proposes a universal prompt-based spatio-temporal predictor that adapts to a wide range of urban forecasting tasks.

The growing use of deep learning and LLM in time series forecasting, which requires massive volumes of data, calls for compression methods that preserve predictive information under edge-device and sensor constraints.

\paragraph{Time Series Data Compression} 

There are two main aspects of time series compression: one along the temporal dimension and the other along the channel dimension. Existing time-series compression methods primarily focus on storage and computational efficiency through temporal reduction, which can be categorized into three types. (1) Representation-based methods \cite{RW1,RW2,RW3} use mathematical functions or learned nonlinear mappings to obtain compact latent representations. (2) Dictionary-based methods \cite{RW6,RW7} exploit structural redundancy via similarity measures or codebook matching. (3) Sequential algorithms \cite{RW4,RW5,RW6,RW7} process data incrementally using information-theoretic principles for online or model-free prediction. On the other hand, general-purpose dimension-reduction techniques—such as Principal Component Analysis (PCA) \cite{Intro1}, Multi-Dimensional Scaling (MDS) \cite{Intro2}, Isomap \cite{Intro3}, t-distributed Stochastic Neighbor Embedding (t-SNE) \cite{Intro4}, and Uniform Manifold Approximation and Projection (UMAP) \cite{Intro5}—are often applied for channel compression without task-specific design. For example, DeepGLO \cite{Intro6} adopt low-rank matrix factorization inspired by the principles of PCA to extract global features, although not specifically for time series data compression.

\paragraph{Multiple-Input Multiple-Output (MIMO)} 

Techniques such as MIMO with Sparse Code Multiple Access \cite{RW13,RW14}, Non-Orthogonal Multiple Access \cite{RW15,RW16}, and Code Division Multiple Access \cite{RW17,RW18} have been proposed to allow different signals to share the same communication channel.
A key principle in their design is to ensure that signals remain distinguishable and do not interfere with each other during transmission, typically through coding, modulation, or power allocation strategies.
However, time series prediction requires preserving temporal context and inter-channel complementary information under compression transmission, which these methods do not support.
Thus, while we are inspired by their channel sharing strategies, they are not directly applicable to predictive tasks.

\section{Problem Definition}

Before presenting the design of our method, we first describe the format and assumptions of the input multichannel time series data. 
We then introduce the objective function that guides our compression and prediction framework.

\paragraph{Definition 1:}
The number of input channels \( C \in \mathbb{N} \) of the input time series data \( X \in \mathbb{R}^{L \times C} \) refers to the input dimension that is distinct from the temporal axis with number of timestamps \(L\).

\paragraph{Assumption 1:}
The multi-channel time series input \( X_{t-L:t}^C \in \mathbb{R}^{L \times C} \) contains latent seasonal patterns, and the input length \(L\) is assumed to be longer than the most dominant seasonal period. Here, “seasonal period” denotes the dominant frequency component (e.g., via FFT), not a strict requirement for an obvious seasonal pattern in the data.

\paragraph{Objective Function}
Motivated by the Cobb–Douglas production function~\cite{Exper1}, we define the objective function as a multiplicative function of prediction accuracy and runtime, as shown below:

{
\setlength{\abovedisplayskip}{-6pt}
\setlength{\belowdisplayskip}{0pt}
\setlength{\abovedisplayshortskip}{-6pt}
\setlength{\belowdisplayshortskip}{0pt}

\begin{equation}
\begin{aligned}
\min_{\theta} \quad
&\mathcal{F}_{\text{pred}}\left(\hat{X}_{t:t+H}^C,\ X_{t:t+H}^C\right) \cdot 
\mathcal{F}_{\text{time}}\left(\theta\right) \\
\text{s.t.} \quad 
&\operatorname{dim}\left(\text{Compressed}(X_{t-L:t}^C)\right) = (L, 1)
\end{aligned}
\label{eq:simple-objective}
\end{equation}
}
In this formulation, \(\theta\) denotes a set of parameters corresponding to the compression, prediction, and decompression modules. The variables \(\hat{X}_{t:t+H}^C\) and \(X_{t:t+H}^C\) represent the predicted and ground truth sequences for the next \(H\) time steps. The term \(\mathcal{F}_{\text{pred}}\) quantifies the prediction error (e.g., Mean Squared Error), while \(\mathcal{F}_{\text{time}}\) represents the total inference latency. The constraint enforces that the compressed representation maintains a single channel, i.e., \(\operatorname{dim} = (L, 1)\).

In this design as shown in Fig. \ref{fig:Method1}, a multichannel time series input \( X_{t-L:t}^C \in \mathbb{R}^{L \times C} \) is first compressed into a single-channel representation \( X_{t-L:t} \in \mathbb{R}^{L \times 1} \). This compressed sequence is used to predict the next \(H\) time steps, resulting in \( X_{t:t+H} \in \mathbb{R}^{H \times 1} \), which is then decompressed to reconstruct the multichannel output \( X_{t:t+H}^C \in \mathbb{R}^{H \times C} \).

\section{Observations and Theoretical Foundations}

In this section, we first identify the redundancy phenomenon in multichannel time series data. Based on this observation, we present Theorem 1, which reveals the opportunity to remove redundant information while retaining predictive information during the compression process. Building on this theoretical opportunity, we introduce Theorem 2 to support that our proposed method is predictability-aware.

\subsection{Redundancy in Multichannel Time Series}

\paragraph{Multi-channel time series data can be represented by a limited number of channels.} As shown in Table~\ref{tab:pca-transposed}, the multichannel time series data across all datasets exhibit substantial redundancy. For example, in the \textit{Sensor Drift} dataset, only 2 out of 321 principal components are sufficient to retain 95\% of the total variance, and the top 50 components capture 100\% of it. Similarly, \textit{NYC taxi} and \textit{DC bike} require fewer than 15 components to preserve 95\% of the variance, with over 99\% already explained by the top 50. Additionally, the first two components alone explain a dominant portion of the variance (e.g., 92.4\% and 5.7\% in \textit{Sensor Drift}), indicating that the data lie near a low-dimensional manifold. These observations validate the presence of strong redundancy in the input channels and motivate the use of compact representations through channel compression.
{
\setlength{\textfloatsep}{3pt}
\setlength{\intextsep}{3pt}
\setlength{\floatsep}{4pt}
\setlength{\abovecaptionskip}{2pt}
\setlength{\belowcaptionskip}{2pt}

\begin{table}[ht]
\centering
\small
\caption{
PCA summary across datasets.
\textit{PCs @95\% var.}: Number of principal components needed to retain 95\% of variance;
\textit{Top 50 var.}: Cumulative variance explained by top 50 PCs;
\textit{PC1/PC2 var.}: Variance explained by the 1st and 2nd principal components.
}
\begin{tabular}{lccc}
\toprule
\textbf{Metric} & \textbf{NYC\_taxi} & \textbf{DC\_bike} & \textbf{Sensor Drift} \\
\midrule
PCs @95\% var. & 15 / 321 & 13 / 321 & 2 / 321 \\
Top 50 var. & 0.9926 & 0.9908 & 1.0000 \\
PC1 var. & 0.5422 & 0.5723 & 0.9242 \\
PC2 var. & 0.1525 & 0.1993 & 0.0569 \\
\bottomrule
\end{tabular}
\label{tab:pca-transposed}
\end{table}
}
{
\setlength{\textfloatsep}{3pt}
\setlength{\intextsep}{3pt}
\setlength{\floatsep}{4pt}
\setlength{\abovecaptionskip}{2pt}
\setlength{\belowcaptionskip}{2pt}
\begin{figure}[ht]
    \centering
    \begin{minipage}[b]{0.32\linewidth}
        \includegraphics[width=\linewidth]{Ana1.pdf}
        \centering \text{\footnotesize (a) NYC\_taxi}
    \end{minipage}
    \hfill
    \begin{minipage}[b]{0.32\linewidth}
        \includegraphics[width=\linewidth]{Ana2.pdf}
        \centering \text{\footnotesize (b) DC\_bike}
    \end{minipage}
    \hfill
    \begin{minipage}[b]{0.32\linewidth}
        \includegraphics[width=\linewidth]{Ana3.pdf}
        \centering \text{\footnotesize (c) Sensor Drift}
    \end{minipage}
    \caption{Heatmaps of raw multichannel time series data. The vertical axis represents 321 channels, and the horizontal axis denotes 321 time steps. Highlighted regions indicate examples of similar patterns.}
    \vspace{-10pt}
    \label{fig:ana1}
\end{figure}
}
\begin{figure*}[!t]
    \centering
    \includegraphics[width=1\linewidth]{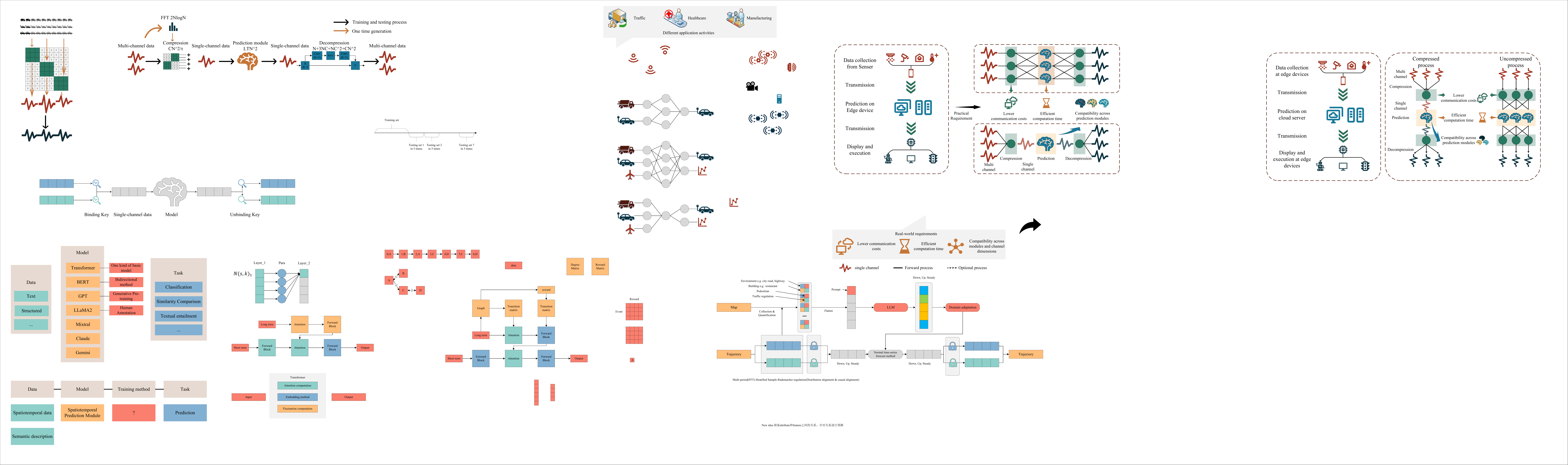}
    \vspace{-20pt}
    \caption{Overall architecture of the compression–decompression framework. Compression and decompression processes take place at the sensor or edge, while prediction is performed in the cloud. The framework is compatible with different deep learning prediction modules, such as transformer-based, MLP-based, and linear-based modules.}
    \vspace{-12pt}
    \label{fig:Method1}
\end{figure*}

\paragraph{Multi-channel time series data exists redundancy about similar pattern and low variation components.}
As shown in Figure~\ref{fig:ana1}, the heatmaps of the raw multichannel time series data reveal clear structural patterns across different datasets. A substantial number of channels exhibit little to no variation over time, as evidenced by large uniform-colored regions. In addition, multiple channels demonstrate similar temporal evolution patterns (such as the red box in Figure \ref{fig:ana1}), suggesting redundancy and high correlation across channels.
For example, in the DC bike dataset, certain areas show very few rentals. However, these data cannot be simply discarded, since there are still occasional rentals at specific times, although rentals are absent for most of the time. In addition, similar patterns can be observed in the DC dataset, which may arise from the same functional zones of the city or from nearby stations.
These observations reveal the existence of compressible structure in data and justify an effective channel compression strategy prior to downstream prediction. To enable tractable modeling of such redundancy, the following assumption is made:

\subsection{Retaining Predictive Information}

\paragraph{Assumption 2: Gaussian redundancy assumption in multichannel time series.}
To model the redundancy or low-variation components often observed in multichannel time series data, we assume that these components follow a zero-mean multivariate Gaussian distribution. Specifically, the redundancy-induced uncertainty across all channels and time steps is represented as:
\begin{equation}
    \varepsilon \sim \mathcal{N}(0,\ \sigma^2 I_{CL}),
\end{equation}
where \( \varepsilon \in \mathbb{R}^{CL} \) is the vectorized noise across all channels and time steps, \( C \) is the number of channels, \( L \) is the sequence length, and \( I_{CL} \) is the identity matrix of size \( CL \times CL \). Each per-channel noise vector \( \varepsilon^{(i)} \in \mathbb{R}^{L} \) is assumed to be independent and identically distributed. 

Motivated by Information Bottleneck (IB) \cite{Method6}, we have the following theorem on multi-channel time series. 

\paragraph{Theorem 1: Upper bound on predictive information reduction under compression}

Under Assumption 2, consider a compressed representation normalized to have unit signal variance. The IB admits the following upper bound:
\begin{equation}
    \mathcal{L}_{\text{IB}} \leq \frac{1}{2} \log \left(1 + \frac{1}{CL \cdot \sigma^2} \right).
\end{equation}
A detailed derivation is provided in the Appendix \ref{sec:upperbound}.

\paragraph{Remark 1}
This result highlights a fundamental trade-off: when the original number of channels \( C \) or the sequence length \( L \) is small, the upper bound on the IB loss becomes larger, indicating a higher risk of losing predictive information under compression. Conversely, larger \( C \) or \( L \) lead to a tighter bound, implying that high-dimensional inputs offer greater potential for safe and effective compression without sacrificing prediction accuracy. This insight theoretically supports the feasibility of compressing multichannel time series in scenarios with abundant channel or temporal resolution.

\subsection{Predictability Awareness Compression}

Circular convolution \cite{Intro17} provides a fixed-width vector operation for binding two representations and has been widely used as a method for compositional encoding in distributed representations. We treat this as a baseline mechanism for compression.

\paragraph{Definition 2: Correlation as a measure of predictability in temporal sequences.}
In time series data, the Pearson correlation coefficient (ranging from 0 to 1) between a historical segment and a future segment reflects the degree of predictability \cite{Intro15, Intro16}, with higher correlation indicating stronger temporal dependency or clearly seasonal pattern. Specifically, for two aligned sequences \( x, y \in \mathbb{R}^L \), a higher Pearson correlation coefficient implies that one sequence is more informative about the other.

\paragraph{Theorem 2: Circular keys preserve single channel predictability information.}

Given Assumption 1 and Definition 2, when the circular key \( K \in \mathbb{R}^L \) shares the same seasonal period as the latent dominant seasonal period of the input sequence \( X_{t-L:t}^C \), the circular convolution can retain the seasonal alignment and thereby preserve predictive information in the compressed representation \( Y \).
{
\setlength{\abovedisplayskip}{2pt}
\setlength{\belowdisplayskip}{0pt}
\setlength{\abovedisplayshortskip}{2pt}
\setlength{\belowdisplayshortskip}{0pt}
\begin{equation}
\begin{split}
    Y^{i} &= \text{Encode}(K^{i}, X^{i}) = X^{i} * K^{i}, \\
    &\text{s.t. } k_t = k_{t \bmod \tau}, \quad \text{for some fixed } \tau \ll L,
\end{split}
\end{equation}
}
where the key enforces a repeating structure with seasonal period \(\tau\). The proof is provided in Appendix  \ref{sec:predictability}.

\paragraph{Remark 2} A formal justification based on the Pearson correlation coefficient shows that when the seasonal key aligns with the latent seasonal period in the input, the encoded representation maximizes predictability, achieving a value of \( R = 1 \).

\paragraph{Remark 3} From a signal processing perspective, the seasonal circular key acts as a matched filter, amplifying frequency components that correspond to the latent seasonality in the input. This alignment ensures that the compressed representation \( Y \) emphasizes task-relevant information while filtering out redundant or irrelevant patterns.

\paragraph{Corollary 1: Inter-channel predictability alignment via Least Common Multiple Seasonal Period (LCMP)}

Given Theorem 2, consider a multichannel input \( \{X^i\}_{i=1}^{C} \), where each channel \( X^i \in \mathbb{R}^L \) exhibits a latent dominant seasonal pattern with seasonal period \( \tau_i \). To preserve inter-channel seasonal consistency and maintain the predictability of the combined representation, we define a shared seasonal period
\begin{equation}
    \tau = \mathrm{lcm}(\tau_1, \tau_2, \dots, \tau_C),
\end{equation}
and constrain each circular key \( K^i \in \mathbb{R}^L \) to satisfy \( k^i_t = k^i_{t \bmod \tau} \). Each channel is then encoded as \( Y^i = \text{Encode}(K^i, X^i) \), and the final compressed representation is the channel-wise summation:
\begin{equation}
    Y = \sum_{i=0}^{C-1} Y^i = \sum_{i=0}^{C-1} \text{Encode}(K, X^i).
\end{equation}

\paragraph{Remark 4}
This seasonal shared-period encoding ensures that all encoded sequences remain aligned, enabling effective cross-channel compression and preserving the overall temporal structure in the latent representation. The seasonal keys \( K \) from different channels can be identical, since they follow the same seasonal period \( \tau \). In the next part, we will introduce a sparse method with reduced compression time, with minor limitations in predictability.

\section{Compression and Decompression Framework}

Based on the theoretical analysis above,
this section illustrates the process of compression and decompression, as well as the time complexity of sparse optimization. 
As shown in Fig. \ref{fig:Method1}, the framework is simple and flexibly compatible with existing time series prediction modules. The compression process has been justified in the former section with two steps (this section introduces a more unified and sparse method), and the following decompression process also involves two error-removal steps. Then, we prove that the framework with a single-channel predictor has lower time complexity than a direct multichannel predictor. Additional details on the pseudocode and training strategy are provided in the Appendix \ref{sec:pseudocode}.

\subsection{Compression}

Based on the former explanation and justification about predictability protection compression, the compression process is shown below. 
A sparse compression strategy is introduced to further reduce the computational burden on edge devices.

Concretely, each channel \( X^i \in \mathbb{R}^L \) is partitioned into non\-/overlapping fragments of length \(\tau\), where \(\tau\) is the least common period (LCP) across all channels. Only complete segments are encoded, and any remaining tail shorter than \(\tau\) is omitted. The compressed output is computed as:
\begin{equation}
    Y = \sum_{i=0}^{C-1} \sum_{j=0}^{M-1} \text{Encode}(X_{j\tau:(j+1)\tau}^i, K),
\end{equation}
where \( M = \left\lfloor \frac{L}{\tau} \right\rfloor \), and \( K \) is the seasonal key shared across all segments. The time complexity of this sparse compression process is reduced from \( \mathcal{O}(CL^2) \) to \(\mathcal{O}\left( \frac{CL^2}{\tau} \right)\). When the period is chosen as \( \tau = \frac{L}{\log L} \), the complexity becomes \(\mathcal{O}(CL \log L)\)

\subsection{Decompression}

The decompression process is designed to mitigate the reconstruction errors introduced by the simplifying assumptions in HRR, particularly the assumption that input data follows a Gaussian distribution—an assumption that often does not hold for real-world time series.

\paragraph{Decoding process via circular correlation.}
Inspired by HRR, the basic decoding process uses circular correlation with seasonal keys, defined as:
\begin{equation}
\begin{aligned}
    \tilde{X}^i &= \text{Decode}(K, \hat{Y}) \\
    &= \left[ \sum_{z=0}^{n-1} \hat{y}_z k_{z},\ 
               \sum_{z=0}^{n-1} \hat{y}_z k_{z-1},\ 
               \dots,\ 
               \sum_{z=0}^{n-1} \hat{y}_z k_{z-h} \right],
\end{aligned}
\end{equation}
where \( \hat{Y} = [\hat{y}_0, \hat{y}_1, \dots, \hat{y}_{h-1}] \) is the predicted representation from the forecasting model, \( h \) denotes the length of the predicted future sequence to be reconstructed, and \( k_z \) is the \( z \)-th element of the decoding kernel \( K \). Under HRR, this yields:
\begin{equation}
\tilde{X}^i[t] \approx (\sum k^2) \cdot \sum_{c=1}^{C} x_t^c + \eta_t,
\end{equation}
where \( \eta_t \) denotes accumulated intra-channel interference due to overlapping shifts in the circular convolution.

\paragraph{Reducing intra-channel errors via orthogonal keys.}
To reduce intra-channel interference, we use orthogonal circular key matrices \( K \), where each key shift is orthogonal to others: \( K^\top K = I \). Such matrices can be efficiently constructed via two FFTs \cite{Method5}. In this setting, different time shifts become mutually orthogonal, nullifying off-timestamp interference in decoding:
{
\setlength{\abovedisplayskip}{1pt}
\setlength{\belowdisplayskip}{0pt}
\setlength{\abovedisplayshortskip}{1pt}
\setlength{\belowdisplayshortskip}{0pt}
\begin{equation}
\tilde{X}^i[t] = (\sum k^2) \cdot \sum_{c=1}^{C} x_t^c
\end{equation}
}

\paragraph{Inter-channel error correction via residuals.}
While orthogonality mitigates intra-channel interference, inter-channel errors still remain. To address this, we introduce a residual correction block. The final decompression is expressed as:
\begin{equation}
\label{eq:decompression}
\hat{X} = \text{Dense}\left( \frac{\text{Copy}(\tilde{X})}{\sum k^2} + \mathrm{Residual}\left( \frac{\text{Copy}(\tilde{X})}{\sum k^2} \right) \right)
\end{equation}
where \( \tilde{X} = \text{Decode}(K, Y) \) with orthogonal circular key matrices, \(\text{Copy}(\cdot)\) duplicates or projects the decoded result into the full channel dimension, \(\mathrm{Residual}(\cdot)\) denotes a lightweight correction block (2 conv layers + ReLU), \(\text{Dense}(\cdot)\) is fully connected layer. This decomposition corrects for shared-channel entanglement and significantly enhances decompression fidelity.

\subsection{Time Complexity Superiority}

This part provides a time complexity analysis for compression and decompression framework.

\paragraph{Theorem 3: Computational Superiority Condition.}

Let \( \mathcal{O}(L)_\text{comp} \) denote the total computational complexity of our compression framework followed by a single-channel predictor, and let \( \mathcal{O}(L)_\text{naive} \) denote that of a naive multichannel predictor, which shares the same structure as the single-channel predictor except that it operates on all channels directly. Here, \( E \) denotes the number of training epochs and \( D \) denotes the depth of predictor. Then:
\begin{equation}
\mathcal{O}(L)_\text{comp} < \mathcal{O}(L)_\text{naive} \\
\Rightarrow 
C > \frac{DE}{DE - 1 - \frac{1}{\tau}}.
\end{equation}
\subsubsection{Corollary 2: Simplified Superiority Condition.}
Under the asymptotic regime \( DE \gg 1 \) and \( \frac{1}{\tau} \to 0 \), the condition reduces to \(C > 1\). The proof of Proposition 4 and Corollary is provided in Appendix \ref{sec:Time complexity analysis}.

\paragraph{Remark 5}
Theorem 3 and Corollary 2 together indicate that the proposed compression-decompression framework offers a computational advantage over the naive multichannel predictor when the number of input channels \( C \) exceeds one. Furthermore, the overhead introduced by compression and decompression becomes subdominant. As a result, the scaling advantage of the proposed framework remains intact in practical high-dimensional time series settings.

\section{Experiments}

In this section, we first introduce the experimental settings, including datasets, baselines, and evaluation metrics. We then analyze the results of the comparative experiments from different metrics aspect.

\begin{table*}[t]
  \centering
  \small
  \caption{Cobb–Douglas Production Index (CDPI) results. Lower values indicate better overall performance. Underlined values denote the best performance in each setting. The number of channels corresponds to the compression ratio, as our framework compresses all channels into one.}\setlength{\tabcolsep}{0.8mm} 
  \vspace{-10pt}
  \begin{tabular}{cl|ccc|ccc|ccc}
    \toprule
    &&\multicolumn{3}{c|}{MLP Based Comparison}&\multicolumn{3}{c|}{Transformer Based Comparison}&\multicolumn{3}{c}{Linear Based Comparison}\\
    \multicolumn{2}{c|}{Dataset \& Channel} 
    & \textbf{PCDF-MLP} & \textbf{TSMixer} & \shortstack{Ratio\\(MLP/TS)} 
    & \textbf{PCDF-Trans} & \textbf{PatchTST} & \shortstack{Ratio\\(Trans/Patch)} 
    & \textbf{PCDF-Linear} & \textbf{HDMixer} & \shortstack{Ratio\\(Linear/HD)} \\
    \midrule
    \multirow{5}*{\shortstack{NYC taxi\\0.1/24}}
    &5  & 0.000459 & \underline{0.000397} & 1.156 & \underline{0.000340} & 0.000716 & 0.475 & \underline{0.000310} & 0.002464 & 0.126 \\
    &10 & \underline{0.000524} & 0.000567 & 0.924 & \underline{0.000386} & 0.001026 & 0.376 & \underline{0.000348} & 0.003966 & 0.088 \\
    &20 & \underline{0.000581} & 0.000735 & 0.790 & \underline{0.000478} & 0.001853 & 0.258 & \underline{0.000968} & 0.006231 & 0.155 \\
    &30 & \underline{0.000664} & 0.000885 & 0.750 & \underline{0.000564} & 0.002532 & 0.223 & \underline{0.000528} & 0.007694 & 0.069 \\
    &40 & \underline{0.000761} & 0.000969 & 0.785 & \underline{0.000745} & 0.003158 & 0.236 & \underline{0.000632} & 0.012362 & 0.051 \\
    \midrule
    \multirow{5}*{\shortstack{DC\_bike\\0.3/30}}
    &5  & \underline{0.000539} & 0.000647 & 0.833 & \underline{0.000413} & 0.001063 & 0.389 & \underline{0.000505} & 0.009265 & 0.055 \\
    &10 & \underline{0.000619} & 0.000864 & 0.716 & \underline{0.000414} & 0.001693 & 0.244 & \underline{0.000366} & 0.016486 & 0.022 \\
    &20 & \underline{0.000691} & 0.001150 & 0.601 & \underline{0.000643} & 0.003208 & 0.200 & \underline{0.000543} & 0.028265 & 0.019 \\
    &30 & \underline{0.000804} & 0.001345 & 0.598 & \underline{0.000949} & 0.003812 & 0.249 & \underline{0.000622} & 0.037110 & 0.017 \\
    &40 & \underline{0.000984} & 0.001500 & 0.656 & \underline{0.001235} & 0.005163 & 0.239 & \underline{0.003392} & 0.050074 & 0.068 \\
    \midrule
    \multirow{5}*{\shortstack{Electricity\\0.1/24}}
    &5  & 0.000299 & \underline{0.000215} & 1.391 & \underline{0.000219} & 0.000337 & 0.650 & \underline{0.000169} & 0.001188 & 0.142 \\
    &10 & \underline{0.000242} & 0.000285 & 0.849 & \underline{0.000234} & 0.000611 & 0.383 & \underline{0.000195} & 0.001984 & 0.098 \\
    &20 & \underline{0.000394} & 0.000448 & 0.879 & \underline{0.000319} & 0.000972 & 0.328 & \underline{0.000335} & 0.004585 & 0.073 \\
    &30 & \underline{0.000430} & 0.000523 & 0.822 & \underline{0.000404} & 0.001315 & 0.307 & \underline{0.000312} & 0.007746 & 0.040 \\
    &40 & \underline{0.000525} & 0.000697 & 0.753 & \underline{0.000534} & 0.001740 & 0.307 & \underline{0.000425} & 0.013574 & 0.031 \\
    \midrule
    \multirow{5}*{\shortstack{Solar Energy\\0.25/24}}
    &5  & \underline{0.000765} & 0.000980 & 0.781 & \underline{0.000538} & 0.001336 & 0.403 & \underline{0.000508} & 0.007004 & 0.073 \\
    &10 & \underline{0.000586} & 0.000915 & 0.641 & \underline{0.000506} & 0.001576 & 0.321 & \underline{0.000416} & 0.007520 & 0.055 \\
    &20 & \underline{0.000631} & 0.001087 & 0.581 & \underline{0.000541} & 0.002316 & 0.234 & \underline{0.000512} & 0.015104 & 0.034 \\
    &30 & \underline{0.000730} & 0.001384 & 0.527 & \underline{0.000625} & 0.003065 & 0.204 & \underline{0.000596} & 0.026226 & 0.023 \\
    &40 & \underline{0.000857} & 0.001401 & 0.612 & \underline{0.000750} & 0.003609 & 0.208 & \underline{0.000869} & 0.039396 & 0.022 \\
    \midrule
    \multirow{5}*{\shortstack{Sensor Drift\\0.1/30}}
    &5  & \underline{0.000548} & 0.001070 & 0.512 & \underline{0.000434} & 0.001573 & 0.276 & \underline{0.000368} & 0.010083 & 0.037 \\
    &10 & \underline{0.000638} & 0.001385 & 0.461 & \underline{0.000456} & 0.002456 & 0.186 & \underline{0.000472} & 0.013823 & 0.034 \\
    &20 & \underline{0.000846} & 0.002084 & 0.406 & \underline{0.000666} & 0.004751 & 0.140 & \underline{0.000792} & 0.024024 & 0.033 \\
    &30 & \underline{0.000951} & 0.002270 & 0.419 & \underline{0.000841} & 0.006407 & 0.131 & \underline{0.000861} & 0.031081 & 0.028 \\
    &40 & \underline{0.001077} & 0.002886 & 0.373 & \underline{0.000929} & 0.009078 & 0.102 & \underline{0.000850} & 0.033141 & 0.026 \\
    \midrule
    \multirow{5}*{\shortstack{Weather\\0.1/25}}
    &5  & \underline{0.000342} & 0.000434 & 0.788 & \underline{0.000273} & 0.000989 & 0.276 & \underline{0.000425} & 0.001812 & 0.235 \\
    &10 & \underline{0.000394} & 0.000547 & 0.720 & \underline{0.000306} & 0.001193 & 0.257 & \underline{0.000267} & 0.003822 & 0.070 \\
    &20 & \underline{0.000569} & 0.000731 & 0.778 & \underline{0.000382} & 0.002106 & 0.182 & \underline{0.000391} & 0.007379 & 0.053 \\
    &30 & \underline{0.000590} & 0.000883 & 0.668 & \underline{0.000535} & 0.003086 & 0.173 & \underline{0.000475} & 0.011419 & 0.042 \\
    &40 & \underline{0.000630} & 0.001081 & 0.583 & \underline{0.000528} & 0.004326 & 0.122 & \underline{0.000533} & 0.022849 & 0.023 \\
    \bottomrule
  \end{tabular}
  \vspace{-5pt}
  \label{tab:Exper1}
\end{table*}

\subsection{Experimental Settings}

\paragraph{Datasets.}
We evaluate our design on datasets from a diverse set of edge-based sensing applications, including (1) traffic: NYC taxi \cite{Exper2} and DC\_bike \cite{Exper3}; 
(2) energy: solar energy generation \cite{Exper4} and electricity consumption \cite{Exper5}; (3) 
manufacturing: gas sensor array drift \cite{Exper7}; and (4) weather \cite{Exper6}.




\paragraph{Baselines and setup.}
We implement our framework with multiple existing predictors (MLP, Transformer, Linear, and Convolution, as detailed in the Appendix).  
Each such combination is referred to as an integration (e.g., PCDF-MLP, PCDF-Trans, PCDF-Linear, PCDF-Conv).  
The architectural details of each single-channel predictor are described as follows:

\textbullet\ \textbf{MLP integration}: Built upon the TSMixer-based MLP, this variant incorporates an additional flatten layer, one linear layer, and a ReLU activation, appended to the end of the original structure.

\textbullet\ \textbf{Transformer integration}: Based on the PatchTST encoder, this variant extends the original architecture by adding a flatten layer, two linear layers, and a ReLU activation.

\textbullet\ \textbf{Linear integration}: A linear predictor derived from HDMixer, consisting of multiple linear layers, with a flatten operation inserted before a final fully connected layer.

\textbullet\ \textbf{Convolution integration}: A convolutional module comprising two convolutional layers, followed by a flatten operation, two fully connected layers, and a ReLU activation function.

We compare these integrations with state-of-the-art models, including PatchTST~\cite{Intro9}, Autoformer~\cite{Intro15}, HDMixer~\cite{Intro12}, DLinear~\cite{Intro10}, MSGNet~\cite{Intro13}, and TSMixer~\cite{Intro11}, to evaluate performance and runtime trade-offs.  
Since, to the best of our knowledge, there is no existing multichannel compression framework specifically designed for time series prediction (as discussed in the Introduction and Related Work), no direct baselines are available for this setting.  
All models follow the same experimental setup, with the number of prediction channels set to \( C \in \{5, 10, 20, 30, 40\} \).  
All experiments are conducted on two NVIDIA RTX 4090 GPUs, each equipped with 24 GB of memory.

\begin{figure*}[!ht]
    \centering

    

    \begin{minipage}[b]{1.0\linewidth}
        \includegraphics[width=\linewidth]{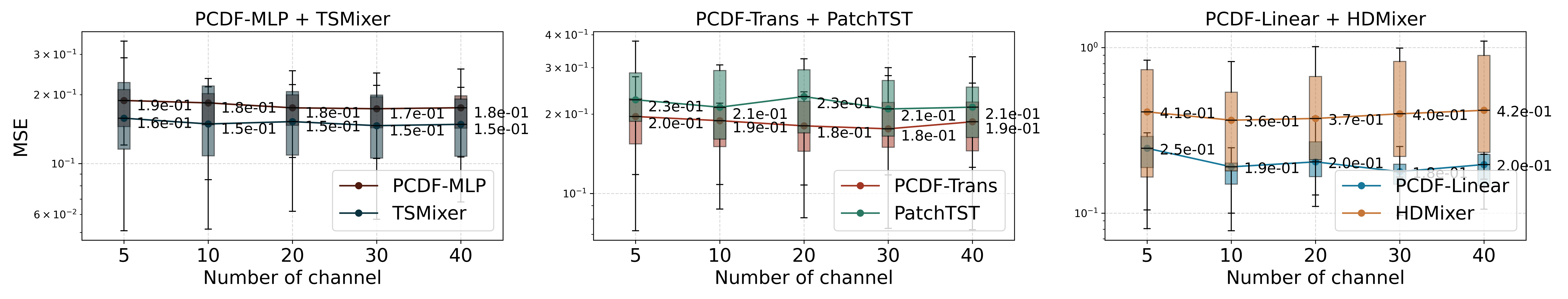}
        \centering \text{\footnotesize (a) MSE}
    \end{minipage}
    
    \begin{minipage}[b]{1.0\linewidth}
        \includegraphics[width=\linewidth]{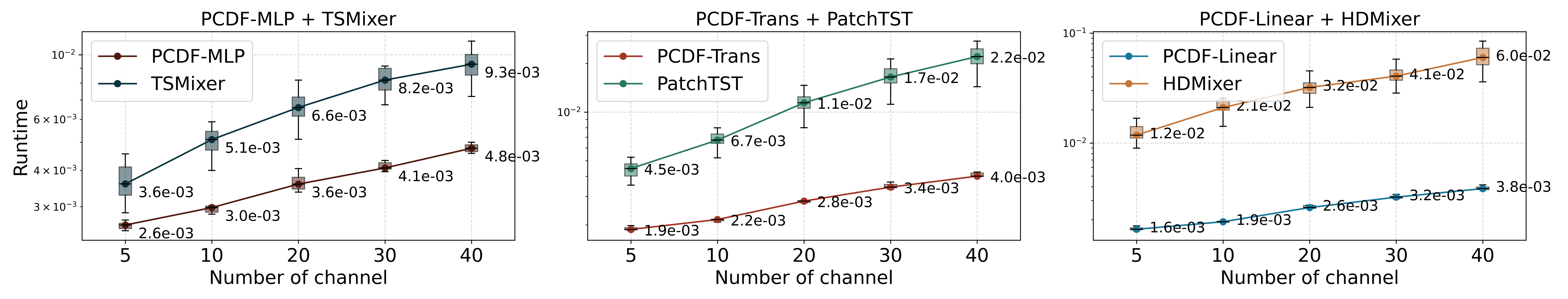}
        \centering \text{\footnotesize (b) Runtime (seconds)}
    \end{minipage}
    \vspace{-10pt}
    \caption{Boxplot of MSE and runtime across channel configurations and datasets}
    \vspace{-15pt}
    \label{fig:Exper1}
\end{figure*}

\paragraph{Metric.}

Evaluation metrics include Mean Squared Error (MSE), runtime, and the Cobb–Douglas Production Index (CDPI). Runtime is obtained from the testing process, not from the training process; therefore, the number of training epochs does not affect the experimental results.

CDPI follows the idea of the Cobb–Douglas production function~\cite{Exper1}, which uses a multiplicative combination to model the joint contribution of multiple factors. We combine MSE and runtime in a similar fashion to capture the trade-off and enable a comprehensive analysis of prediction accuracy and computational efficiency.
\begin{equation}
\label{eq:Cobb–Douglas production index}
    \text{CDPI} = \text{MSE} \cdot \text{Runtime}.
\end{equation}
In our experiments, we do not apply normalization when computing CDPI, in order to preserve the raw scale of MSE and runtime. Normalizing either metric before multiplication could introduce bias, as it implicitly assigns weights and may distort the true impact of each component.

\paragraph{Training.}

In this experiment, compression and decompression are trained on a single device. For real deployment, one option is centralized training with distributed inference, meaning that the model is trained in the cloud and then the model or parameters are transmitted to edge or sensor devices for compression and decompression. Another option is distributed training with distributed inference, where models are trained on different sensors or edge devices and both compression and decompression are performed locally on those devices.

\subsection{CDPI Analysis}

Based on the multiply of MSE and runtime results for all models (see details in Appendix \ref{sec:Result analysis with MSE} and \ref{sec:Result analysis with run time}), CPDI results are presented in Table~\ref{tab:Exper1}. The compression–decompression integrations achieve lower combined index values compared to the representative baseline models across different datasets in most cases. These results suggest that the compression–decompression framework has the potential to achieve better overall performance by reducing prediction time and lowering prediction errors with the high compression ratio.

From the perspective of different datasets, those with clear seasonal patterns, such as Electricity, tend to exhibit lower combined index values. This indicates that such datasets either require less computational time or result in lower prediction errors after processing, leading to better overall performance. Conversely, datasets with less pronounced seasonal patterns tend to have higher combined index values, reflecting increased prediction errors or longer computational times.

\begin{table}[htbp]
\centering
\small
\vspace{-5pt}
\caption{\small Training time (in seconds) under different train horizons}
\vspace{-10pt}
\setlength{\tabcolsep}{3pt}
\begin{tabular}{lcccc}
\hline
\textbf{Method} & \textbf{channel=5} & \textbf{channel=10} & \textbf{channel=20} & \textbf{channel=40} \\
\hline
PCDF-MLP & 0.17080 & 0.14992 & 0.17723  & 0.58769 \\
PCDF-Trans & 0.25910 & 0.24673 & 0.27429 & 0.67120 \\
PCDF-Linear   & 0.16091 & 0.17340 & 0.20060 & 0.48645 \\
PatchTST    & 0.63854 & 0.92027 & 1.47746 & 3.64563 \\
AutoFormer  & 1.53149 & 1.52916 & 1.48094 & 2.48258 \\
HDMixer     & 0.88682 & 1.29958 & 2.06925 & 4.35996 \\
Dlinear     & 0.07111 & 0.12328 & 0.22287 & 0.80155 \\
\hline
\end{tabular}
\label{tab:train_horizon_compaison}
\vspace{-15pt}
\end{table}

\subsection{Runtime and Accuracy Analysis}

Figure~\ref{fig:Exper1} shows the variance of MSE. The integrations exhibit lower variance across different datasets compared to their corresponding base models, indicating that our framework provides better stability for the underlying predictors.

In terms of overall MSE and runtime values, the integrations generally achieve lower values than their base models, except for the PCDP-MLP and its corresponding base model TSMixer in the MSE metric. This suggests that our framework can effectively reduce runtime, decrease MSE, and improve overall performance compared to other methods built upon the same predictors, despite the exception observed with TSMixer.

{
\setlength{\textfloatsep}{3pt}
\setlength{\intextsep}{3pt}
\setlength{\floatsep}{2pt}
\setlength{\abovecaptionskip}{4pt}
\setlength{\belowcaptionskip}{2pt}
\begin{table}[ht]
  \caption{Compression-decompression framework ablation study via MSE (values rounded to 4 decimal places). Abbreviations: Comp. = Compression, Decomp. = Decompression, Decode. = Decoder, Encode. = Encoder.}
  \vspace{-2pt}
  \label{tab:ablation}
  \centering
  \resizebox{\linewidth}{!}{%
  \begin{tabular}{l|ccccc}
    \toprule
    Method & NYC taxi & DC bike & Electricity & Solar Energy & Sensor Drift \\
    \midrule
    Comp.-Decomp. & 0.1800 & \textbf{0.2153} & \textbf{0.1073} & \textbf{0.2172} & \textbf{0.2389} \\
    Rand.-Decomp. & 0.1983 & 0.2595 & 14.8035 & 18.5596 & 0.2533 \\
    Comp.-Decode. & 0.2132 & 1257.9678 & 0.2435 & 9.8324 & 0.5709 \\
    Encode.-Decomp. & 0.3232 & 5.4924 & 143.5187 & 1.9980 & 12.7464 \\
    Encode.-Decode. & \textbf{0.1546} & 1.5364 & 0.1737 & 1.6343 & 0.3028 \\
    \bottomrule
  \end{tabular}
  }
  \vspace{-5pt}
\end{table}
}

\subsection{Training Time Analysis}
Table \ref{tab:train_horizon_compaison} shows the training time of the proposed framework under the same number of epochs. The PCDF-family scales well with channel count, with slower growth in training time—PCDF-Trans (×2.6), PCDF-Linear (×3.0), and PCDF-MLP (×3.4)—compared to PatchTST (×5.7), HDMixer (×4.9), and DLinear (×11.3). Even with 40 channels, all PCDF variants finish training in less than one second, showing clear speed-ups over methods without the compression–decompression framework. These results indicate that the proposed methods become more efficient when the number of input channels increases.

\subsection{Ablation Analysis}


In the ablation study, a classic encoder–decoder framework is used to replace the compression and decompression components in order to justify the effectiveness of the proposed framework. In addition, a random key is used in place of the seasonal key to evaluate the effectiveness of the compression component guided by seasonal information. Therefore, the ablation study includes four variants: (1) random-key compression–decompression, (2) compression–decoder, (3) encoder–decompression, and (4) encoder–decoder.

In the random-key variant, the results vary significantly across datasets, suggesting that the random key fails to uncover the underlying predictive patterns in some datasets. This highlights the importance of using the seasonal key within the compression\-/decompression framework.

For the compression decoder and encoder decompression variants, since compression and decompression are co-designed and interdependent, using only one of the components leads to unstable results. This indicates that optimal performance relies on the synergy between both compression and decompression modules.

In the Encoder–Decoder variant, although lower MSE is observed on the NYC Taxi dataset, the performance remains unstable overall. This may be attributed to the fact that the encoder cannot effectively preserve or extract predictive patterns, and the corresponding decoder may fail to recover the single-channel signal in a lossless manner. The inferior and unstable results of these ablation variants validate the superiority of our framework in both extracting latent predictive information during compression and performing accurate reconstruction during decompression.

\section{Conclusion}

This paper proposes a novel predictability-aware compression\-/decompression framework tailored for multichannel time series prediction. By leveraging a seasonality-guided convolution process with circularly orthogonal key matrices, the framework effectively preserves latent temporal dependencies during compression and mitigates reconstruction errors during decompression. Extensive experiments across diverse datasets and prediction backbones demonstrate that our framework consistently achieves lower runtime and improved prediction accuracy, as evidenced by reduced Cobb–Douglas production indices. Overall, this work offers a flexible and efficient solution for compressing multichannel time series data without sacrificing predictive performance, paving the way for more resource-efficient forecasting in edge–cloud environments.

\paragraph{Limitations and future work.} Despite strong model compatibility, the framework requires retraining when applied to new domains, limiting its out-of-domain generalization. Additionally, under extreme compression ratios, training instability such as gradient explosion may occur. While this is alleviated via scale-invariant gradient clipping, we attribute the root cause to the large dimensional disparity between compressed and decompressed representations. Future work will explore principled mechanisms to bridge this gap and enhance training robustness.


\begin{acks}
We acknowledge the support of NSF (2520547). Any opinions, findings, and conclusions or recommendations expressed in this material are those of the authors and do not necessarily reflect the views of the NSF.
\end{acks}

\bibliographystyle{ACM-Reference-Format}
\bibliography{sample-base}

\appendix

\section{Appendix}

\subsection{Upper Bound Derivation of the Information Bottleneck Objective}
\label{sec:upperbound}

We adopt the Information Bottleneck (IB) principle~\cite{Method6}, with the objective formulated as:
\vspace{-1pt}
\begin{equation}
\mathcal{L}_{\text{IB}} = I(Y;\ X^C_{t:t-L}) - \beta I(Y;\ X^C_{t:t+H}),
\end{equation}
where \( X^C_{t:t-L} \in \mathbb{R}^{C \times L} \) is the historical input,  
\( X^C_{t:t+H} \in \mathbb{R}^{C \times H} \) is the future target to predict,  
\( Y \in \mathbb{R}^{L} \) is the compressed representation derived from \( X^C_{t:t-L} \),  
and \( \beta > 0 \) is a trade-off parameter balancing compression and prediction.

\subsubsection*{Upper Bound on \( I(Y; X^C_{t:t-L}) \)}

Following Assumption 2, we assume the historical input includes Gaussian redundancy:
\vspace{-1pt}
\begin{equation}
X^C_{t:t-L} = S + \varepsilon,\quad \varepsilon \sim \mathcal{N}(0, \sigma^2 I_{CL}),
\end{equation}
where \( S \in \mathbb{R}^{C \times L} \) is the true signal and \( \varepsilon \in \mathbb{R}^{CL} \) is i.i.d. noise.  
We apply a linear compression:
\vspace{-1pt}
\begin{equation}
Y = K X^C_{t:t-L} = K S + K \varepsilon,
\end{equation}
where \( K \in \mathbb{R}^{L \times CL} \) is the projection matrix.  
Assuming a Gaussian encoder, the mutual information upper bound becomes:
\vspace{-1pt}
\begin{equation}
I(Y;\ X^C_{t:t-L}) \leq \frac{1}{2} \log \left(1 + \frac{1}{CL \cdot \sigma^2} \right).
\end{equation}

\subsubsection*{Lower Bound on \( I(Y;\ X^C_{t:t+H}) \)}

We now bound the prediction term using a variational approximation \( q(x \mid y) \) of the true conditional:
\vspace{-1pt}
\begin{align}
I(Y;\ X^C_{t:t+H}) &= \mathbb{E}_{p(y,\hat{x})} \left[ \log q(\hat{x} \mid y) \right] 
+  \notag \\
&\underbrace{\mathbb{E}_{p(y,\hat{x})} \left[ \log \frac{p(\hat{x} \mid y)}{q(\hat{x} \mid y)} \right]}_{\text{KL divergence} \geq 0} + H(X^C_{t:t+H})
\end{align}
Hence, the variational lower bound is:
\vspace{-1pt}
\begin{equation}
I(Y;\ X^C_{t:t+H}) \geq \mathbb{E}_{p(y,\hat{x})} \left[ \log q(\hat{x} \mid y) \right] + H(X^C_{t:t+H}).
\end{equation}
If the model is well trained, \( q(\hat{x} \mid y) \approx p(\hat{x} \mid y) \), and:
\vspace{-1pt}
\begin{equation}
I(Y;\ X^C_{t:t+H}) \approx H(\hat{X}^C_{t:t+H}) - H(X^C_{t:t+H}) = 0.
\end{equation}

\subsubsection*{Final Upper Bound}

Combining both bounds, we obtain:
\vspace{-1pt}
\begin{equation}
\mathcal{L}_{\text{IB}} \leq \frac{1}{2} \log \left(1 + \frac{1}{CL \cdot \sigma^2} \right)
\end{equation}

This shows that increasing channel count \( C \), history length \( L \), or noise variance \( \sigma^2 \) will decrease the IB objective, promoting stronger compression.

\subsection{Predictability Preservation in Compression}
\label{sec:predictability}

We provide a formal justification for how circular convolution preserves temporal predictability when the key structure aligns with the latent seasonality of the input.

Let each channel \( X^i \in \mathbb{R}^L \) be partitioned into \( m \) non-overlapping segments of length \( \tau \), such that \( L = m\tau \). Let the key \( K \in \mathbb{R}^{\tau} \) be a seasonal vector with period \( \tau \), extended to length \( L \) via circular repetition.

Each compressed channel is encoded as:

\begin{equation}
\begin{aligned}
    Y^i[t] &= \text{Encode}(K, X^i)[t] = \sum_{z=0}^{L-1} x^i_z \cdot k_{(t - z) \bmod L}, \\
& t \in \{0, 1, \dots, L-1\}.
\end{aligned}
\end{equation}

We can express the first \( \tau \) outputs compactly as:
\begin{equation}
Y^i_{0:\tau} = 
\begin{bmatrix}
K^{\langle 0 \rangle} \odot X^i \\
K^{\langle 1 \rangle} \odot X^i \\
\vdots \\
K^{\langle \tau-1 \rangle} \odot X^i
\end{bmatrix},
\end{equation}
where \( K^{\langle t \rangle} \) denotes the circular shift of \( K \) by \( t \) positions, and \( \odot \) denotes element-wise multiplication followed by summation.

To assess the preservation of seasonal structure, we compute the Pearson correlation coefficient between \( Y^i[t] \) and its delayed version \( Y^i[t+\tau] \):
\vspace{-1pt}
\begin{equation}
\begin{aligned}
&R = \\
&\frac{
\sum_{t=0}^{\tau-1} \left(Y^i[t] - \bar{Y}^i_{0:\tau}\right)
              \left(Y^i[t+\tau] - \bar{Y}^i_{\tau:2\tau}\right)
}{
\sqrt{ \sum_{t=0}^{\tau-1} \left(Y^i[t] - \bar{Y}^i_{0:\tau}\right)^2 }
\cdot
\sqrt{ \sum_{t=0}^{\tau-1} \left(Y^i[t+\tau] - \bar{Y}^i_{\tau:2\tau}\right)^2 }
},
\end{aligned}
\end{equation}
where \( \bar{Y}^i_{0:\tau} \) and \(\bar{Y}^i_{\tau:2\tau} \) denote the mean values of the first and second seasonal segments of the encoded signal \( Y^i \), respectively. This formulation evaluates the seasonal correlation between two adjacent cycles in the compressed output. Then \( K^{\langle t \rangle} \) = \( K^{\langle t+\tau \rangle} \):
\vspace{-1pt}
\begin{equation}
R = \frac{\sum_{t=0}^{\tau-1} (Y^i[t] - \bar{Y}^i)^2}{\sqrt{\sum (Y^i[t] - \bar{Y}^i)^2} \cdot \sqrt{\sum (Y^i[t] - \bar{Y}^i)^2}} = 1.
\end{equation}

Hence, when the circular key aligns with the latent seasonal period \( \tau \), the circular convolution operation effectively preserves temporal dependencies in the encoded output, ensuring the retention of predictive information.

\begin{table*}[ht]
  \caption{MSE comparison across models}
  \label{tab:whole_mse}
  \centering
  \small
  \vspace{-10pt}
  \resizebox{\textwidth}{!}{ 
  \begin{tabular}{cl|cccc|cccccc}
    \toprule
    \multicolumn{2}{c|}{Dataset \& Channel} & PCDF-MLP & PCDF-Conv & PCDF-Trans & PCDF-Linear & PatchTST & Autoformer & HDMixer & Dlinear & MSGNET & TSMixer \\
    \midrule
    \multirow{5}*{\shortstack{NYC taxi\\0.1/24}}
    &5 & \textbf{0.17001} & 0.17123 & 0.17185 & 0.17617 & 0.18207 & 0.22242 & 0.21746 & 0.22952 & 0.17786 & \underline{0.11583} \\
    &10 & \textbf{0.17345} & 0.17402 & 0.17866 & 0.18100 & 0.16052 & 0.21227 & 0.19901 & 0.22148 & 0.18730 & \underline{0.10712} \\
    &20 & \textbf{0.16690} & 0.16755 & 0.17004 & 0.38246 & 0.17685 & 0.20932 & 0.20868 & 0.23443 & 0.15630 & \underline{0.10621} \\
    &30 & \textbf{0.16273} & 0.18081 & 0.16454 & 0.16447 & 0.16722 & 0.19818 & 0.20644 & 0.23384 & 0.16985 & \underline{0.10325} \\
    &40 & 0.16375 & 0.17950 & 0.18567 & \textbf{0.16808} & 0.15947 & 0.20778 & 0.21135 & 0.23004 & 0.16840 & \underline{0.10497} \\
    \midrule
    \multirow{5}*{\shortstack{DC bike\\0.3/30}}
    &5 & \textbf{0.20727} & 0.21891 & 0.21993 & 0.30601 & 0.24833 & 0.25229 & 0.84002 & 0.33392 & 0.28390 & \underline{0.19984} \\
    &10 & 0.20484 & 0.32077 & \textbf{0.19908} & 0.19993 & 0.26378 & 0.25559 & 0.82432 & 0.35534 & 0.24590 & \underline{0.18705} \\
    &20 & 0.20499 & \textbf{0.20131} & 0.23476 & 0.20643 & 0.29622 & 0.26901 & 1.01525 & 0.34106 & 0.24141 & \underline{0.18827} \\
    &30 & 0.19953 & 0.19884 & 0.27986 & \textbf{0.19940} & 0.25127 & 0.24671 & 0.99303 & 0.32456 & 0.24285 & \underline{0.17986} \\
    &40 & \textbf{0.20115} & 0.20429 & 0.26227 & 0.81534 & 0.25409 & 0.29385 & 1.01694 & 0.33351 & 0.24126 & \underline{0.18116} \\
    \midrule
    \multirow{5}*{\shortstack{Electricity\\0.1/24}}
    &5 & 0.12057 & 0.11257 & 0.11797 & \textbf{0.10462} & 0.07222 & 0.13478 & 0.08058 & \underline{0.06247} & 0.06887 & 0.05089 \\
    &10 & \textbf{0.08506} & 0.10490 & 0.10813 & 0.10000 & 0.08733 & 0.13689 & 0.07827 & \underline{0.05899} & 0.06863 & 0.05172 \\
    &20 & 0.10636 & \textbf{0.10162} & 0.10774 & 0.10971 & 0.08091 & 0.14275 & 0.12867 & \underline{0.06489} & 0.07619 & 0.06189 \\
    &30 & 0.10527 & \textbf{0.09370} & 0.11765 & 0.09699 & 0.07376 & 0.14071 & 0.16501 & \underline{0.06934} & 0.07419 & 0.05711 \\
    &40 & 0.10686 & 0.11419 & 0.12568 & \textbf{0.10561} & 0.07290 & 0.14487 & 0.16060 & \underline{0.07126} & 0.07742 & 0.06797 \\
    \midrule
    \multirow{5}*{\shortstack{Solar energy\\0.25/24}}
    &5 & \textbf{0.28963} & 0.34007 & \underline{0.27720} & 0.29853 & 0.37834 & 0.32659 & 0.78084 & 0.41998 & 0.30821 & 0.34265 \\
    &10 & \underline{0.19477} & \textbf{0.19823} & 0.22016 & 0.20110 & 0.30252 & 0.26677 & 0.52998 & 0.32197 & 0.23241 & 0.22869 \\
    &20 & \underline{0.18351} & 0.20605 & \textbf{0.19106} & 0.20172 & 0.28950 & 0.28373 & 0.71582 & 0.30441 & 0.22477 & 0.21239 \\
    &30 & \underline{0.18445} & \textbf{0.19313} & 0.18724 & 0.19172 & 0.27392 & 0.27739 & 0.92214 & 0.28723 & 0.21932 & 0.20562 \\
    &40 & 0.18703 & \underline{0.18488} & \textbf{0.18851} & 0.22520 & 0.25131 & 0.28551 & 1.09463 & 0.27789 & 0.21166 & 0.19491 \\
    \midrule
    \multirow{5}*{\shortstack{Sensor drift\\0.1/30}}
    &5 & \underline{0.21170} & 0.24100 & 0.23216 & \textbf{0.22586} & 0.29965 & 0.29230 & 0.60090 & 0.49438 & 0.26063 & 0.23470 \\
    &10 & 0.21635 & 0.22153 & \underline{0.21592} & \textbf{0.24843} & 0.30743 & 0.28553 & 0.54017 & 0.49369 & 0.25783 & 0.23546 \\
    &20 & \underline{0.22143} & 0.26097 & \textbf{0.24296} & 0.29123 & 0.32431 & 0.30663 & 0.53079 & 0.52499 & 0.26535 & 0.25440 \\
    &30 & \underline{0.21964} & 0.31027 & \textbf{0.23310} & 0.25249 & 0.30008 & 0.29319 & 0.53616 & 0.57539 & 0.26360 & 0.24865 \\
    &40 & \underline{0.21537} & 0.25694 & \textbf{0.23342} & 0.22679 & 0.32976 & 0.29579 & 0.53791 & 0.57626 & 0.26061 & 0.25885 \\
    \midrule
    \multirow{5}*{\shortstack{Weather\\0.1/25}}
    &5 & 0.13686 & \textbf{0.13677} & 0.14842 & 0.26743 & 0.20510 & 0.20085 & 0.14801 & 0.15422 & 0.19025 & \underline{0.11550} \\
    &10 & 0.13915 & 0.19315 & \textbf{0.14120} & \textbf{0.13915} & 0.16099 & 0.18119 & 0.17340 & 0.15355 & 0.16385 & \underline{0.11092} \\
    &20 & 0.14017 & 0.14220 & \underline{0.13612} & \textbf{0.15410} & 0.16739 & 0.18602 & 0.21575 & 0.17785 & 0.16017 & 0.11679 \\
    &30 & \textbf{0.13686} & \underline{0.13609} & 0.14488 & 0.14433 & 0.16433 & 0.19715 & 0.26048 & 0.16975 & 0.16483 & 0.11303 \\
    &40 & 0.13583 & 0.13694 & \underline{0.13162} & \textbf{0.13908} & 0.17373 & 0.19966 & 0.29841 & 0.16809 & 0.16277 & 0.11571 \\
    \bottomrule
  \end{tabular}
  }
  \vspace{-5pt}
\end{table*}

\begin{table*}[ht]
  \caption{Runtime comparison across models. The reported computational time is measured in seconds}
  \label{tab:whole runtime}
  \centering
  \small
  \vspace{-10pt}
  \resizebox{\textwidth}{!}{
  \begin{tabular}{cl|cccc|cccccc}
    \toprule
    \multicolumn{2}{c|}{Dataset \& Channel} & PCDF-MLP & PCDF-Conv & PCDF-Trans & PCDF-Linear & PatchTST & Autoformer & HDMixer & Dlinear & MSGNET & TSMixer \\
    \midrule
    \multirow{5}*{\shortstack{NYC taxi\\0.1/24}}
    &5  & 0.00270 & 0.00152 & \textbf{0.00198} & 0.00176 & 0.00393 & 0.02552 & 0.01133 & \underline{0.00047} & 0.06311 & 0.00343 \\
    &10 & 0.00302 & 0.00173 & \textbf{0.00216} & 0.00192 & 0.00639 & 0.02513 & 0.01993 & \underline{0.00077} & 0.07779 & 0.00529 \\
    &20 & 0.00348 & 0.00235 & \textbf{0.00281} & 0.00253 & 0.01048 & 0.02604 & 0.02986 & \underline{0.00140} & 0.10720 & 0.00692 \\
    &30 & 0.00408 & 0.00301 & \textbf{0.00343} & 0.00321 & 0.01514 & 0.02641 & 0.03727 & \underline{0.00197} & 0.12478 & 0.00857 \\
    &40 & 0.00465 & 0.00359 & \textbf{0.00401} & 0.00376 & 0.01980 & 0.02738 & 0.05849 & \underline{0.00246} & 0.19804 & 0.00923 \\
    \midrule
    \multirow{5}*{\shortstack{DC bike\\0.3/30}}
    &5  & 0.00260 & 0.00155 & \textbf{0.00188} & 0.00165 & 0.00428 & 0.02676 & 0.01103 & \underline{0.00046} & 0.05958 & 0.00324 \\
    &10 & 0.00302 & 0.00183 & \textbf{0.00208} & 0.00183 & 0.00642 & 0.02688 & 0.02000 & \underline{0.00079} & 0.07809 & 0.00462 \\
    &20 & 0.00337 & 0.00234 & \textbf{0.00274} & 0.00263 & 0.01083 & 0.02796 & 0.02784 & \underline{0.00135} & 0.11492 & 0.00611 \\
    &30 & 0.00403 & 0.00293 & \textbf{0.00339} & 0.00312 & 0.01517 & 0.02711 & 0.03737 & \underline{0.00198} & 0.11755 & 0.00748 \\
    &40 & 0.00489 & 0.00356 & \textbf{0.00471} & 0.00416 & 0.02032 & 0.02742 & 0.04924 & \underline{0.00254} & 0.15987 & 0.00828 \\
    \midrule
    \multirow{5}*{\shortstack{Electricity\\0.1/24}}
    &5  & 0.00248 & 0.00144 & \textbf{0.00186} & 0.00162 & 0.00466 & 0.02887 & 0.01474 & \underline{0.00047} & 0.07086 & 0.00423 \\
    &10 & 0.00285 & 0.00175 & \textbf{0.00216} & 0.00195 & 0.00700 & 0.02988 & 0.02535 & \underline{0.00085} & 0.09254 & 0.00551 \\
    &20 & 0.00370 & 0.00254 & \textbf{0.00296} & 0.00305 & 0.01201 & 0.03053 & 0.03563 & \underline{0.00147} & 0.14381 & 0.00724 \\
    &30 & 0.00408 & 0.00303 & \textbf{0.00343} & 0.00322 & 0.01783 & 0.03135 & 0.04694 & \underline{0.00213} & 0.15304 & 0.00915 \\
    &40 & 0.00491 & 0.00383 & \textbf{0.00425} & 0.00402 & 0.02387 & 0.03307 & 0.08452 & \underline{0.00278} & 0.29734 & 0.01026 \\
    \midrule
    \multirow{5}*{\shortstack{Solar energy\\0.25/24}}
    &5  & 0.00264 & 0.00151 & \textbf{0.00194} & 0.00170 & 0.00353 & 0.02359 & 0.00897 & \underline{0.00040} & 0.04768 & 0.00286 \\
    &10 & 0.00301 & 0.00185 & \textbf{0.00230} & 0.00207 & 0.00521 & 0.02305 & 0.01419 & \underline{0.00071} & 0.06095 & 0.00400 \\
    &20 & 0.00344 & 0.00237 & \textbf{0.00283} & 0.00254 & 0.00800 & 0.02271 & 0.02110 & \underline{0.00126} & 0.08181 & 0.00512 \\
    &30 & 0.00396 & 0.00292 & \textbf{0.00334} & 0.00311 & 0.01119 & 0.02281 & 0.02844 & \underline{0.00175} & 0.08308 & 0.00673 \\
    &40 & 0.00458 & 0.00363 & \textbf{0.00398} & 0.00386 & 0.01436 & 0.02294 & 0.03599 & \underline{0.00215} & 0.10797 & 0.00719 \\
    \midrule
    \multirow{5}*{\shortstack{Sensor drift\\0.1/30}}
    &5  & 0.00259 & \textbf{0.00250} & 0.00187 & 0.00163 & 0.00525 & 0.03950 & 0.01678 & \underline{0.00053} & 0.08043 & 0.00456 \\
    &10 & 0.00295 & \textbf{0.00298} & 0.00211 & 0.00190 & 0.00799 & 0.04868 & 0.02559 & \underline{0.00111} & 0.10722 & 0.00588 \\
    &20 & 0.00382 & \textbf{0.00413} & 0.00274 & 0.00272 & 0.01465 & 0.05048 & 0.04526 & \underline{0.00155} & 0.17748 & 0.00819 \\
    &30 & 0.00433 & \textbf{0.00476} & 0.00361 & 0.00342 & 0.02135 & 0.04070 & 0.05797 & \underline{0.00231} & 0.16854 & 0.00913 \\
    &40 & 0.00500 & \textbf{0.00505} & 0.00398 & 0.00375 & 0.02753 & 0.03732 & 0.06161 & \underline{0.00285} & 0.22166 & 0.01115 \\
    \midrule
    \multirow{5}*{\shortstack{Weather\\0.1/25}}
    &5  & 0.00250 & 0.00138 & \textbf{0.00184} & 0.00159 & 0.00482 & 0.03028 & 0.01224 & \underline{0.00050} & 0.07227 & 0.00376 \\
    &10 & 0.00283 & 0.00172 & \textbf{0.00217} & 0.00192 & 0.00741 & 0.03045 & 0.02204 & \underline{0.00082} & 0.09493 & 0.00493 \\
    &20 & 0.00406 & 0.00238 & \textbf{0.00281} & 0.00254 & 0.01258 & 0.02922 & 0.03420 & \underline{0.00152} & 0.14586 & 0.00626 \\
    &30 & 0.00431 & 0.00324 & \textbf{0.00369} & 0.00329 & 0.01878 & 0.03096 & 0.04384 & \underline{0.00210} & 0.20596 & 0.00781 \\
    &40 & 0.00464 & 0.00357 & \textbf{0.00401} & 0.00383 & 0.02490 & 0.03088 & 0.07657 & \underline{0.00288} & 0.28855 & 0.00934 \\
    \bottomrule
  \end{tabular}
  }
  \vspace{-5pt}
\end{table*}

\subsection{Time Complexity Justification}
\label{sec:Time complexity analysis}

The time complexity of the proposed framework is derived by considering the complete pipeline, including \textbf{compression}, \textbf{decompression}, and \textbf{prediction}. The total computational complexity is given by:
\vspace{-1pt}
\begin{equation}
\begin{aligned}
    \mathcal{O}(L)_\text{comp} 
    &= \underbrace{DEL^2}_{\text{Prediction}} 
    + \underbrace{\frac{CL^2}{\tau} + 2L\log L}_{\text{Compression}} \\
    &\quad + \underbrace{L + 3CL + CL^2 + CL^2}_{\text{Decompression}} \\
    &\approx DEL^2 + \frac{CL^2}{\tau} + CL^2,
\end{aligned}
\end{equation}
where \(E\) denotes the number of training epochs, \(L\) is the number of timestamps, \(C\) \((C < L)\) is the number of channels, \(D\) is the depth of the predictor, and \(\tau\) is the least common multiple (LCM) seasonal period across different channels. In the asymptotic approximation, we neglect lower-order terms.

The time complexity of the decompression module, dominated by residual blocks, can be broken down into standard neural network operations:
\vspace{-1pt}
\begin{equation}
\label{eq:decompressiontime}
\begin{aligned}
\mathcal{O}(L)_\text{decompression} 
&= L + 3CL + CL^2 + CL^2
\end{aligned}
\end{equation}

In contrast, a naive multichannel predictor that processes all channels jointly without compression incurs the following computational complexity:
\vspace{-1pt}
\begin{equation}
    \mathcal{O}(L)_\text{naive} = DECL^2
\end{equation}

To determine the minimum number of channels \(C\) for which the proposed method achieves lower computational cost than the naive baseline, we compare both complexities:
\vspace{-1pt}
\begin{equation}
\begin{aligned}
    \mathcal{O}(L)_\text{naive} &> \mathcal{O}(L)_\text{comp} \\
    DECL^2 &> DEL^2 + CL^2 + \frac{CL^2}{\tau} \\
    \Rightarrow \quad C &> \frac{DE}{DE - 1 - \frac{1}{\tau}}
\end{aligned}
\end{equation}

Under the assumptions that \(DE \gg 1\) and \(\frac{1}{\tau} \to 0\), the lower bound simplifies to:
\vspace{-1pt}
\begin{equation}
    C > 1
\end{equation}

This result indicates that the proposed compression-based framework offers a computational advantage when the number of channels exceeds one. Furthermore, since the additional costs of compression and decompression are subdominant in the asymptotic regime, they do not alter the scaling benefit of the proposed method.

\subsection{Framework Pipeline with Pseudocode}
\label{sec:pseudocode}

\paragraph{Pseudocode for training process.} 
The pseudocode in Algorithm~\ref{alg:train-infer} describes the training phase of our compression–prediction\-/decompression framework. 

\begin{algorithm}
\caption{Training Phase of Compression–Prediction Framework}
\label{alg:train-infer}
\begin{algorithmic}[1]
\FOR{epoch $= 1$ to $T$}
    \STATE \textbf{Compression:}
    \FOR{$i = 1$ to $C$}
        \STATE Split $X^i \in \mathbb{R}^L$ into segments of length $\tau$: $X^i_{0:\tau}, X^i_{\tau:2\tau}, \dots$
        \FOR{each complete segment}
            \STATE $Y^i \gets \text{Encode}(X^i_{j\tau:(j+1)\tau}, K)$
        \ENDFOR
    \ENDFOR
    \STATE $Y \gets \sum_{i=1}^C Y^i$

    \STATE \textbf{Prediction:}
    \STATE $\hat{Y} \gets \text{Predict}(Y)$

    \STATE \textbf{Decompression:}
    \STATE $\tilde{X} \gets \text{Decode}(K, \hat{Y})$
    \STATE $\hat{X} \gets \text{Dense}\left( \frac{\text{Copy}(\tilde{X})}{\sum k^2} 
    + \text{Residual}\left( \frac{\text{Copy}(\tilde{X})}{\sum k^2} \right) \right)$

    \STATE \textbf{Loss Computation:}
    \STATE Compute $\mathcal{L}$ as defined in Eq.~\ref{eq:loss}

    \STATE \textbf{Parameter Update:}
    \STATE $\theta \gets \theta - \eta \cdot \nabla_\theta \mathcal{L}$
\ENDFOR
\end{algorithmic}
\end{algorithm}

\paragraph{Additional techniques for training the model}

Following Dish-TS~\cite{Method2}, we apply normalization after compression and denormalization before decompression.





\paragraph{Training loss functions}

The total training objective consists of three terms:
\vspace{-1pt}
\begin{equation}
\label{eq:loss}
\begin{aligned}
\mathcal{L} =\ & \underbrace{\|\hat{X} - X\|^2}_{\text{Prediction Loss}} 
+ \alpha \underbrace{\left( \sum_{i=1}^{C} |\text{Residual}_i| - \sum_{i=1}^{C} \hat{X}_i \right)^2}_{\text{Regulation Term}} \\
& + \beta \underbrace{\|\varphi_Y - \varphi_{\hat{Y}}\|^2}_{\text{Latent Consistency}}.
\end{aligned}
\end{equation}

Here, \(\text{Residual}_i\) represents the residual correction output for the \(i\)-th channel, \( \varphi_Y \) and \( \varphi_{\hat{Y}} \) denote the mean values of the original and predicted compressed representations, respectively. This latent consistency term encourages the predictor to preserve global statistics (e.g., mean trends) in the compressed space, ensuring more coherent reconstructions after decompression. The regulation term encourages sparsity and suppresses overfitting.

To improve numerical stability, we use scale-invariant gradient clipping~\cite{Method3}. For each parameter \( w \) and its gradient \( g \), we clip the gradient as:
\vspace{-1pt}
\begin{equation}
    g \leftarrow g \cdot \min\left(1,\ \frac{\alpha \|w\|}{\|g\|} \right),
\end{equation}

where \( \alpha > 0 \) is a hyperparameter, \(g\) denotes the gradient and \(w\) is the corresponding model parameter. 


\subsection{Result Analysis with MSE}
\label{sec:Result analysis with MSE}


As shown in Table \ref{tab:whole_mse}, the best scores are underlined and the second-best are bolded. The PCDF variants (MLP, Conv, Trans, and Linear) frequently achieve top results across datasets. 

\subsection{Result Analysis with Runtime}
\label{sec:Result analysis with run time}

As shown in Table \ref{tab:whole runtime}, our variants (PCDF-MLP, PCDF-Conv, PCDF-Trans, PCDF-Linear) consistently achieve the lowest or second-lowest runtime across datasets and channel settings. These results highlight both the accuracy and efficiency of our approach, especially in resource-limited settings.


\end{document}